\def\year{2020}\relax
\documentclass[letterpaper]{article} 
\usepackage{aaai20}  
\usepackage{times}  
\usepackage{helvet} 
\usepackage{courier}  
\usepackage[hyphens]{url}  
\usepackage{graphicx} 

\usepackage[absolute]{textpos}
\usepackage{graphicx}  
\usepackage{subcaption}
\title{Towards Automated Swimming Analytics Using Deep Neural Networks}
\author{Timothy Woinoski, Alon Harell, and 
\Large \textbf{Ivan V. Baji\'{c}} \\
School of Engineering Science, Simon Fraser University\\ 
8888 University Dr, Burnaby, British Columbia, Canada \\
twoinosk@sfu.ca, aharell@sfu.ca, ibajic@ensc.sfu.ca 
}

\begin{document}

\begin{textblock}{15}(3,1.75)
\centering
\noindent\Large AITS 2020 : The AAAI-20 Workshop on AI in Team Sports, New York, Feb 7 2020 
\end{textblock}

\maketitle

\begin{abstract}
Methods for creating a system to automate the collection of swimming analytics on a pool-wide scale are considered in this paper. There has not been much work on swimmer tracking or the creation of a swimmer database for machine learning purposes. Consequently, methods for collecting swimmer data from videos of swim competitions are explored and analyzed. The result is a guide to the creation of a comprehensive collection of swimming data suitable for training swimmer detection and tracking systems. With this database in place, systems can then be created to automate the collection of swimming analytics.  
\end{abstract}

\section{Introduction}

Swimming analytics are valuable for both swimmers and coaches in improving short and long term performance. They provide measurable methods of quantifying improvement and ability over the course of a swimmers career. Such data can be used at the individual level, when making training plans to improve performance in places where data indicates weakness. It can also be used on the team level. Every swimmer's career comes down to making some kind of a team. Such teams range from simply making a collegiate team to making the country's Olympic National Team. Each team will have coaches whose job is to pick the best swimmers to maximize the team's performance. At all levels, the coaches seek the best swimmers for the present and the future. These decisions are never easy, as all teams have limited space and specific needs in terms of being competitive with other teams. Except in the case of relays, swimming is often perceived an individual sport. But it is important to recognize that it is also a team sport, especially with the addition of the International Swim League (ISL), the world's first league where swimmers compete against other swimmers as a team, much like in the NBA, NHL, and NFL.

\begin{figure}[!t]
\centering
\includegraphics[width=0.45\textwidth]{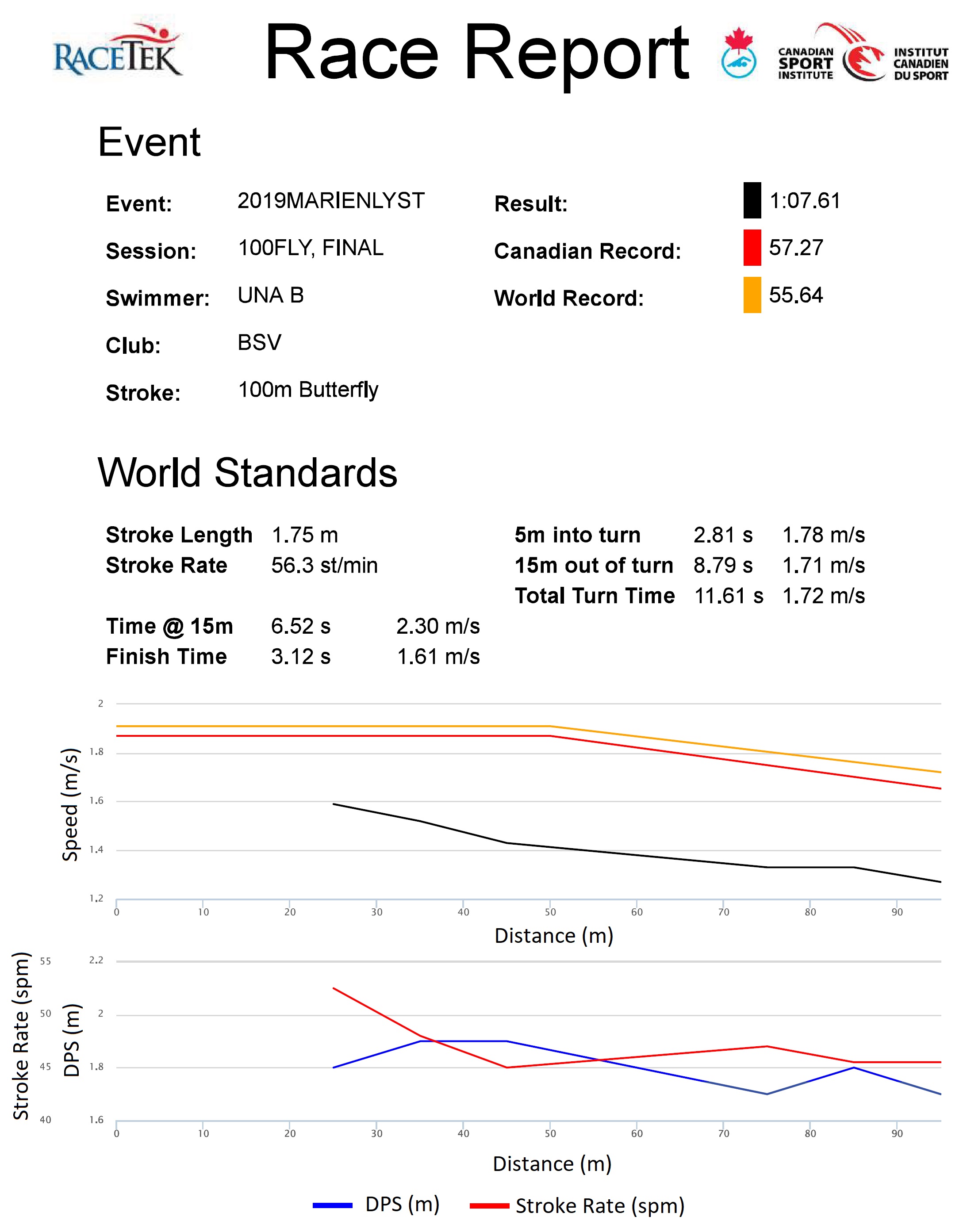}
\caption{A modified example report from RaceTek \cite{raceteck}, found under Video Race Analysis (VRA)}
\label{fig:racetek}
\end{figure}

The motivation for this work is to help produce a system that will automate the collection of swimming analytics in swim competition videos across all participants, using image-based processing methods and tracking algorithms. This would save coaches and athletes many hours analyzing post race videos manually. There is plenty of useful swimming data collected every day. For example, RaceTek \cite{raceteck} provides Canadian swimmers with racing data, as seen in Figure~\ref{fig:racetek}, for all major and some minor competitions; RaceTek has been doing so for many years now. There are other organizations that provide similar services in practice environments, such as Form Swim and TritonWare~\cite{form_swim,tritonwear}. These businesses have been very successful over the past few years. 

The basic services provided by RaceTek are stroke rate analysis, swim velocity analysis and turn analysis collected from videos recorded at swim competitions. All calculations are done by hand using video footage to analyze swimming. Theses tasks are time-consuming and can be automated. The goal of this paper is to discuss the path towards building an automated swimming analytics engine to provide useful  data to coaches and swimmers. Specifically, we address the need to create a dataset for training swimmer detection and tracking models, which could then support automatic generation of currently used analytics~\cite{victor}, as well as create new ones.

\section{Previous Work}
Much work has been done to collect swimming analytics in the training environment. For example,~\cite{zecha} used computer vision methods to estimate swimmer stroke counts and stroke rates on footage collected from a swim channel. The channel was used to control their environment; in addition, they also collected their footage underwater. Thus, in their case, only one swimmer is in the video at any given time. This, however, would be impractical in case of multiple swimmers in a competition. Another example is analytics with accelerometers placed on the body of the swimmers~\cite{Mooney,tritonwear,form_swim}. For the task of collecting swimming analytics across all races, wearable devices are too impractical due to weight, the water dynamics of wearable devices, and the regulations regrading swimming with such devices. 

This leaves analyzing overhead race video of all swimmers as the most widely-available and least intrusive way to collect swimming analytics. Little published research on automated swimming analytics of competition video is available. Some of the main contributions are~\cite{sha,sha_understanding,victor,victor_new}. In~\cite{sha}, the authors used a complex assortment of tracking algorithms to determine the current state of a swimmer and thus the best way to locate their current position. The main problem with their work was that, while it did produce good results, the results were only valid for the one pool their work was dedicated to. If such a system were to be employed in a different pool, it would have to be recalibrated. A more recent work~\cite{victor} gives the most practical results for collecting swimming analytics in a competition environment. Unfortunately, this work can only collect analytics using a video tracking a single swimmer, which means additional detection and video processing is needed to extract such segments from the common multi-swimmer video. The task of counting strokes relies heavily on proper tracking~\cite{victor_new}. In the above works, the main challenge is finding swimmers in a multi-swimmer video scene and tracking their movement using one or just a few camera sources. This is made difficult as swimmers start on a block, dive into the water, turn on a wall, and swim under the water.  

There are few works on tracking humans in aquatic environments, the main ones being~\cite{sha,Chan}. In~\cite{Chan}, the authors performed detection in an aquatic environments using dense optical flow motion map and intensity information. These works use background subtraction to determine the location of swimmers based on a model of the aquatic environment. The methods of swimmer detection proposed in these papers are valid candidates to use to assist in tracking. However, they do not generalize well, as they are heavily dependent on the pools for which they were developed. For their proposed methods to be used at other pools would require recalibration and re-estimating background parameters. 

Fortunately, there is plenty of work on general object tracking in diverse environments, such as~\cite{kalal}. This work assumed that nothing was known about what was going to be tracked and notes that when some prior information is available about objects to be tracked, it is possible to build a more accurate and robust tracker. Due to the very structured nature of swimming competitions, a fairly accurate tracker seems possible. An example of another tracking system using this principle is~\cite{alvar}, which is built upon the You Only Look Once (YOLO) detection system~\cite{Redmon} to aid their tracking algorithm. 
Unfortunately, training such systems requires annotated datasets of swimmers in swimming environments, which currently do not seem to be openly available.

Collecting annotated data is well researched. The creators of the PASCAL VOC challenge~\cite{everingham} spent time creating a well-defined procedure for collecting useful data from online footage. In summary, VOC strove to collect images that accurately represented all possible instances of objects to be detected. This means that all examples of objects have adequate variance in terms of object size, orientation, pose, illumination, position and occlusion. The data also needs to be annotated in a consistent, accurate and exhaustive manner. The VOC challenge used images from the website Flickr, which allowed for classes to be created fulfilling the variance requirements noted. No such set of picture examples exist for swimmers in a racing environment. Fortunately, there is an abundance of race video footage which, when processed, can be used as examples. This race footage can be found mainly on YouTube, as well as through private organizations such as the Canadian Sports Institute or USA Swimming Productions. Currently, there seems to be no open dataset of annotated swimmers.  

\section{Swim Race Footage Variability}
In this section we provide an in-depth look at the pool environment and what one can expect when observing footage of swim competitions. It is important to know such information in order to collect a proper dataset of swimmers,  because such knowledge aids in the collection of scenarios with sufficiently large variability of conditions. There are three main contributors to variance when it comes to capturing swimming races in pool environments. These contributors are: venue, camera angle, and swimming itself. 

\subsection{Venue}
The venue refers to the place of competition, the pool in which the racing occurs. Each venue, or even the same venue, can vary according to many factors, which include the course of the pool, the number of lanes, the lighting, architecture, lane ropes, and the presence of flags.

The \textit{course of a pool} refers to the distance the swimmers must travel in order to complete one pool length. For example, in a long course meter (LCM) pool, swimmers must complete fifty meters before they encounter a wall. There are three main pool courses used competitively in swimming around the world: LCM, short course meters (SCM), and short course yards (SCY). The bulk of SCY competitions are held in the USA. The course of a pool has a big effect on where cameras are placed. In an SCM or SCY race, one camera can easily capture an entire race. In an LCM race, however, the pool is long enough to require multiple cameras to avoid reducing the relative size of the swimmers to a very small fraction of the frame. Another point worth considering is that one venue can host all three courses depending on how the pool is built. Less than optimal performance could be achieved if a model were to be trained on one configuration and then be tested on another, even in the same pool, without other similar training data to support it.

Different pools can have different numbers of \textit{lanes}. Typically, there is an even number of lanes, ranging from six to ten. However, in some situations, competitions can be held in LCM pools but raced as SCY or SCM. This can result in up to a twenty-lane race. Causing even more confusion, some competitions have swimmers racing in one half of the pool, but warming up in the other half. An example would be having lanes 0 to 9 with swimmers racing while having lanes 10 through 19 open to swimmers who are not racing. 

Pool \textit{lighting} is usually kept as constant as possible in high-level competitions to avoid blinding, as swimmers must swim on their front and on their back. Competitions can be indoors or outdoors, resulting in a wide variety of lighting sources and conditions. In competitions where the pool is indoors, the lighting usually comes from the roof, the ceilings are high, and the lighting is even across an entire pool, but each pool can be more or less illuminated than others. In outdoor pools during daytime, light generally comes from the sun and this can result in very bright reflections and very drastic shadows. At nighttime in an outdoor pool, lighting comes from underwater and from above, with varying degrees of illumination. This results in three important factors to consider: illumination, shadow, and glare.

Another important consideration is pool \textit{architecture}, which includes the depth of the pool, what is it made from, the presence of bulkheads, the style of blocks, any markings that may be on the pool bottom, and deck space. Pool depth will affect the relative position of the observed markings at the bottom of the pool. Lane markings are always present and run down the center of each lane. Pools can be made with tile, metal, or other synthetic materials. The composition of materials is less relevant, but the way they absorb and reflect light will affect glare, lighting, and hue in the pool image. Bulkheads are floating walls in the pool that can be moved to change the course of the pool. They look different from the end of the pool mainly because there can be water on the other side of a bulkhead. There are many different types of blocks and they can look drastically different. This is important because a machine learning model could learn that a swimmer is ``on-blocks'' by learning the structure of the blocks rather than the position of the swimmer. 

Besides lane markers, which are generally present in all competition pools, there may be other marks on the bottom of pools. For example, at TYR competitions there are large TYR logos at the bottom of the pool. Such big markings can make swimmers hard to identify when they are underwater. Lastly, one must consider the amount of deck space available in the pool and how it is being utilized. A pool with little deck space and many swimmers walking around it will result in many occlusions of the swimmers in the closest lanes, while a pool with more deck space and fewer people on deck will have fewer occlusions. 

\textit{Lane ropes} are the division between lanes. They run parallel to the pool edges and stop a swimmer from getting too close to another swimmer while also dampening waves. On the world stage there are very strict rules on how lane ropes may look and what color pattern they must adhere to. In general competition however, pools will use whatever lanes ropes are available and thus their detailed structures are of little use for swimmer detection. They are useful in a broad sense as they are fairly straight and define areas where swimmers can and can not be. In a training set, it would be desirable to have a wide variety of examples of pools with different styles of lane ropes.

\textit{Flags} are objects that are always above the swimmers in a competition setting. They are a very consistent source of occlusions in swim competition. The flags indicate to the swimmers when the wall is approaching; they are generally five meters or yards away from each end of the pool and span the width of the pool. In more prestigious competitions, when races that do not require flags occur, the flags are removed for better viewing of the swimmers. Thus, it is important to produce examples of races with and without flags in a training set.

This concludes the list of sources of venue variance. While not all sources of variance may be captured in a given training set, it is desirable to be aware of their existence when dealing with specific pool examples.

\subsection{Camera angle}
When collecting stroke data, the most useful angles are the ones where the swimmers are racing in a direction perpendicular to the view of the camera. A Cartesian coordinate system can be applied to the pool to describe the location of cameras and thus the camera angles that can be accomplished. The x-axis will be defined as the closest wall in the view of the camera such that the swimmers are swimming parallel to it. The z-axis of this coordinate system is parallel to the vertical direction relative to the pool, and the y-axis is parallel to the direction of the blocks or the wall that swimmers turn on.     

In general, the position of the camera along the z-axis is usually at the \textit{pool level} or the \textit{viewing level}. Pool level is roughly the height of a standing person on deck and viewing level is a height at which all swimmers can be put in view by the camera. Ideally, all races are recorded at viewing level. In media designed for the viewer, most races have multiple cameras and thus,  races are recorded with both pool-level footage and view-level footage. The advantage to having pool-level footage is that quick and complex movements such as dives can be more easily captured, but the disadvantage is not having a good view of all swimmers. The advantage of view-level footage is that all swimmers can be seen, but swimmers appear smaller in the scene and so it is more difficult to see what a swimmer is doing.

The camera position along the x-axis, i.e., along the length of the pool,  usually captures one of the following three views: (1) the \textit{dive view}, which means the camera is anywhere before the ten-meter mark closest to the blocks; (2) the \textit{turn view}, which means the camera is past the ten-meter mark of the turn end of the race; and (3) a \textit{mid-pool view}, which is anywhere between the dive view and the turn view. Typically, three main (x,z) camera positions are used: pool-level at dive view, viewing-level at mid-pool view, and pool-level at turn view. It is easier to find footage for these three typical combinations than for the others. This is important when creating a training set. All typical combinations of camera positions must be included for a model to properly generalize the locations of swimmers in a race environment. 

\subsection{Swimming}
What the swimmers are doing in the water is another variable to consider in swimming footage. A model will need to generalize how swimmers look in a pool when they are racing and so it will need examples of all circumstances. There are four different strokes and twenty-nine different races in swimming, not including the different flavors of race when comparing LCM and SCM. Many races are subsets of other races and so each race need not be exclusively collected to get a good representation of that pool. Both genders take part in the sport of swimming and thus variance can come from the different genders. Women race with more suit material than men and so in high definition footage women look different than men in the water. As swimmers age they tend to increase in size and strength and thus a swimmer who is very young will look drastically different in the water than someone who is older. In summary, it is important that all four strokes are included in training video, a roughly even distribution of male and female swimmers, and an even distribution of age groups of swimmers with different speeds.

\subsection{Summary}
We summarize this section by presenting an exhaustive list of sources of variance that should be considered when building a data set of swim race video footage.

\subsubsection{Adequate Variety of Venues} 
\begin{itemize}
    \item Pool lighting, i.e. big shadows or good lighting
    \item Pools with and without bulkheads 
    \item Pools with different block styles
    \item Pools with different depths
    \item Pool courses (LCM, SCM, SCY)
    \item Lanes, including unusual examples
\end{itemize}

The more combinations and variations of these items across the entire dataset, the better. As already mentioned, the same venue can have different pool courses and lane numbers at different competitions. Hence, we propose collecting a variety of footage for each venue as follows.

\subsubsection{Variety for Each Venue} 
A collection of all of the following variants is required across the entire dataset. 
\begin{itemize}
    \item Presence/absence of occlusions from people and flags
    \item Variety of camera angles, especially the three typical camera positions mentioned
    \item Swimming stroke and race distance
    \item Gender and age
\end{itemize}

For each venue it is important that an example of all four stokes are included and races with sprints (roughly thirty to sixty second races) to distance (roughly eight to fifteen minute races) are collected. Each venue should have data captured with all available camera angles. For a given venue, there is no need to have examples from all of the twenty-nine different races with each gender and age group, so long as a variety of ages and genders is present across the entire dataset. It is also not imperative that every competition moves the flags.

\subsubsection{Footage Augmentation}
Some features created by pool venue and camera angle can be effectively simulated by augmenting the dataset. Such suggested augmentations are as follows. 
\begin{itemize}
    \item Flipping images across the y-axis
    \item Shearing or perspective transformation of images to simulate different camera angles
    \item Changing overall brightness, hue and contrast
\end{itemize}

\section{Proposed Method}
We now describe our proposed method of data collection and the creation of a dataset of swimmers racing. As outlined in~\cite{everingham}, it is important that every labelled swimmer is labelled consistently, so that a machine learning algorithm can learn effectively.

\subsection{Obtaining Examples of Swimmers}
The type of data collected is very important to creating a reliable detection system; Ideally there is an adequate variance of situations described in the section above. There are many potential sources for collecting this data, such as \cite{swim_usa,Aust_Institute_Sport,Can_Institute_Sport}. Swim USA’s YouTube page alone has roughly five-hundred races published on it a year. As a first step, examples of swimmers were collected from race footage taken from only one pool; that being the 2019 TYR Pro Swim Series - Bloomington, posted on Youtube. A bigger dataset of pools and races from different sources is more desirable it was decided that before more work was done, some preliminary work would be completed to understand how a model would preform with one pool.

\subsection{Classes of Swimmers}
Referring to the paper \cite{sha}, six classes of swimmers will be collected when creating a swimmer dataset. These classes are as follows: on-blocks, diving, underwater, swimming, turning, and finishing. These six classes, if detected correctly, are valuable in automated collection of swimming metrics such as dive reaction speed, distance off the wall, as well collecting splits, race times, and more. Thus, these six classes will be used in the creation of a dataset. There are points in a race when a swimmer must transition from one class to the next, these transitions must be well defined so that annotations are  collected consistently. 

The following is a guide on how to choose the swimmers class in a frame, given knowledge of the race and the distance completed in the race. This will not include illegal transitions such as underwater to turn.

\begin{enumerate}
    \item \textbf{``On-blocks''}
        \begin{itemize}
            \item ``On-blocks'' is always first
            \item Class starts at the point when the swimmers are on the blocks at the start of the race 
            \item The transition to the next class will be defined as the point when the swimmer is no longer touching the blocks
        \end{itemize}
    \item \textbf{Diving}
        \begin{itemize}
            \item Diving is always after ``On-blocks''
            \item Defined as the point when the swimmer is in mid air and not on the block or underwater or swimming yet
            \item The transition out of diving will be defined as the point when the entire swimmer becomes occluded by the water and splash of the dive entry, in the case that a swimmer fails to completely submerge themselves, skip the underwater class completely and start annotating as swimming
        \end{itemize}
    \item \textbf{Underwater}
        \begin{itemize}
            \item Underwater can only happen after a turn or diving 
            \item Defined as any point in the race when the swimmer is completely submerged, not touching a wall and not swimming
            \item The transition out of underwater will be defined as the point when the swimmer breaks the water with any part of their body to start swimming
            \item Don’t annotate a swimmer if they can’t be seen, i.e. 90\% of swimmer is hidden due to angle, lane ropes and refraction of water
        \end{itemize}
    \item \textbf{Swimming}
        \begin{itemize}
            \item Swimming comes after underwater, diving or turning
            \item Defined as any point in the race when the swimmer is completing legal stroke cycles and not touching a wall
            \item The transition out of swimming into turning can occur on a touch turn or on a flip turn. 
            \item When preforming a touch turn, turning commences when the swimmer touches the wall
            \item When preforming a flip turn, the turn commences when the swimmer is on their front and the head is submerged due to the the flip
            \item The transition out of swimming into finishing is when the swimmer touches the wall and the races has concluded
        \end{itemize}
    \item \textbf{Turning}
        \begin{itemize}
            \item Turning only happens after swimming
            \item Defined as any point in a race when the swimmer comes to a stop near enough to a wall in order to touch or push off the wall
            \item The transition out of turning to underwater is when the swimmer’s feet or possibly, last body part leaves the wall
            \item The point when a swimmer is completely straight can also signify the transition to underwater
            \item In the case that a swimmer fails to completely submerge themselves after a touch, skip the underwater class completely and start annotating as swimming
            
            \item There should be no point at which a turn should not be boxed unless it is cut off by the camera or camera angle as the swimmer is somewhere on the wall
        \end{itemize}
    \item \textbf{Finish}
        \begin{itemize}
            \item Finishing only happens after swimming
            \item Defined as any point after the conclusion of the race distance
            \item Finishing is always the final class of swimmer
        \end{itemize}
\end{enumerate}

\subsection{Class Annotation Details}
This section outlines how to assign a bounding box to each example of a swimmer. In general the box must be the smallest possible box containing the entire swimmer, "Except where the bounding box would have to be made excessively large to include a few additional pixels ($\leq$5\%)" \cite{Annotation_Guidelines}. If 80\% - 90\% of a swimmer is cut off by the camera, do not give them a box. Put a box around a swimmer that can be identified in any way, unless it is cut off by the camera or camera angle. Because there are a variety of situations where this statement becomes ambiguous there will be some general guidelines for specific classes.

\begin{figure*}[ht!]
    \begin{subfigure}[t]{0.3\textwidth}
        \centering
        \includegraphics[height=1.2in]{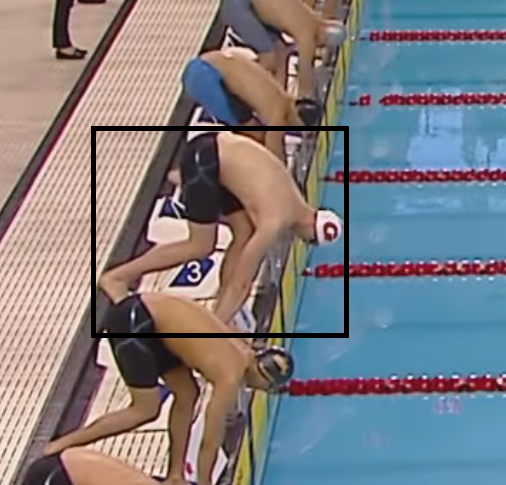}
        \captionsetup{justification=centering}
        \caption[center]{Swimmers on blocks}
        
        \label{fig:on_blocks}
    \end{subfigure}
    ~
    \begin{subfigure}[t]{0.3\textwidth}
        \centering
        \includegraphics[height=1.2in]{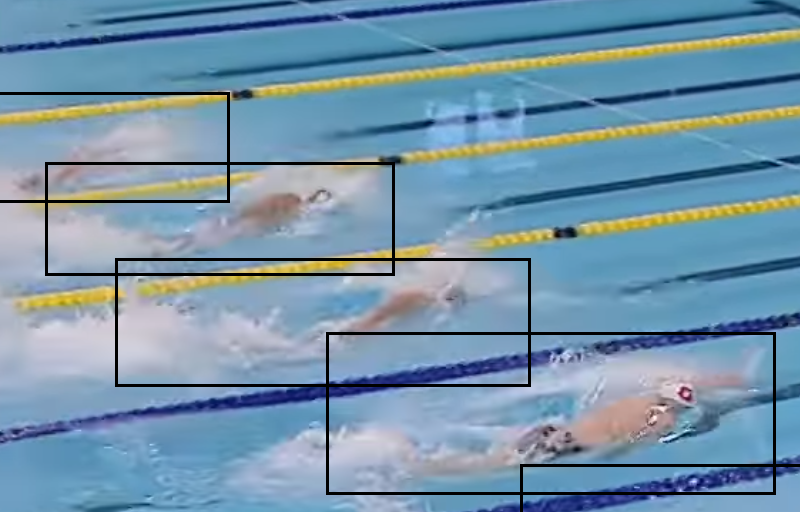}
        \captionsetup{justification=centering}
        \caption{Swimmers swimming}
        \label{fig:swimming}
    \end{subfigure}
    ~
    \begin{subfigure}[t]{0.3\textwidth}
        \centering
        \includegraphics[height=1.2In]{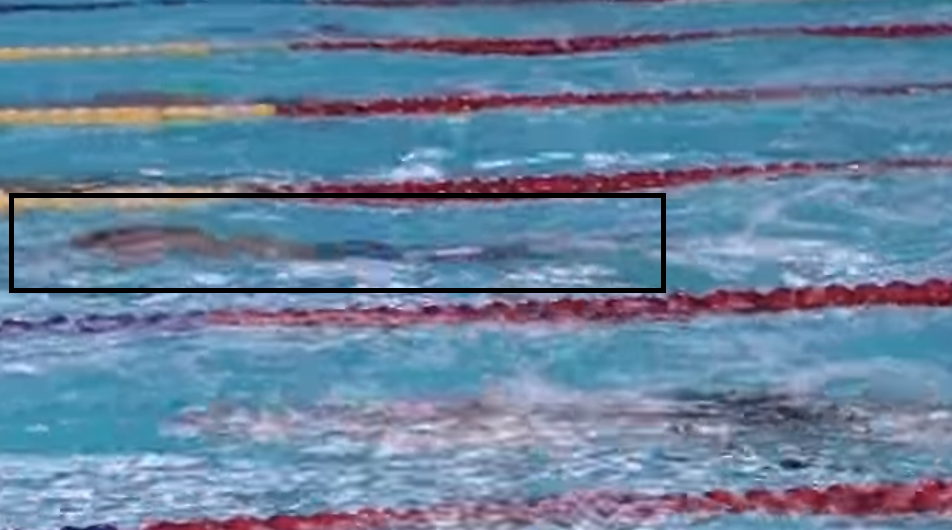}
        \captionsetup{justification=centering}
        \caption{Swimmers underwater}
        \label{fig:underwater}
        \end{subfigure}  
    
    \caption{Examples of swimmers states}
    
\end{figure*}


\subsubsection{``On-blocks''}
For swimmers in the farther lanes and behind other swimmers, add the tightest box possible around all visible parts of the swimmer. If the tip of a foot is visible from behind another swimmer, for example, do not make the box excessively larger than the majority of the swimmer visible. An example would be the swimmer above the annotated swimmer in figure \ref{fig:on_blocks}.  
 

\subsubsection{Swimming}
Stretch the box to include arms and feet. Center the end of the box with the swimmers feet around the splash produced by the kick if the feet are not visible.


\subsubsection{Underwater}
When a swimmer is visible, create the smallest possible box that encompasses the swimmer, see figure \ref{fig:underwater}. When a swimmer becomes too difficult to box accurately do not annotate the swimmer, see the top three swimmers in figure \ref{fig:underwater}. 


\subsubsection{Turning} 
The smallest box shall be made around the swimmer such that it encompasses the swimmer, for all swimmers, regardless of how visible the swimmer is in terms of occlusions. If more than ninety percent of the swimmer is out of camera view then do not annotate the swimmer. 
 

\subsubsection{Finishing}
As a swimmer finishes, they generally look to the clock to see their time. As this happens, they transition from being in horizontal body position, to being in a vertical body position. Due to the refraction of water and bubbles formed by the swimmer, the body of the swimmer becomes invisible to the camera. Thus, a minimal box around what is viable is all that is required.   
 

\subsubsection{Diving}
It can be difficult to determine exactly which swimmer is being annotated. This is because the minimal box including the entirety of one swimmer could also require that the swimmer below is included. Create a minimal box around the swimmer being annotated even if this means the box created also includes a large portion of the other swimmer.

\begin{figure*}[ht!]
    \begin{subfigure}[t]{0.3\textwidth}
        \centering
        \includegraphics[height=1.2in]{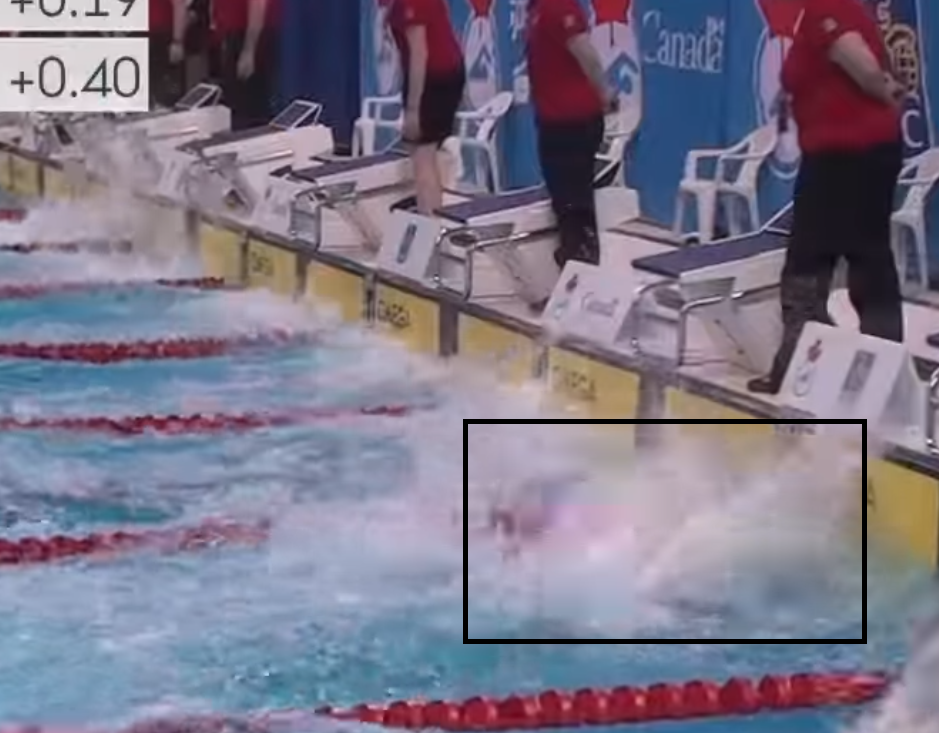}
        \captionsetup{justification=centering}
        \caption[center]{Swimmers turning}
        
        \label{fig:on_blocks}
    \end{subfigure}
    ~
    \begin{subfigure}[t]{0.3\textwidth}
        \centering
        \includegraphics[height=1.2in]{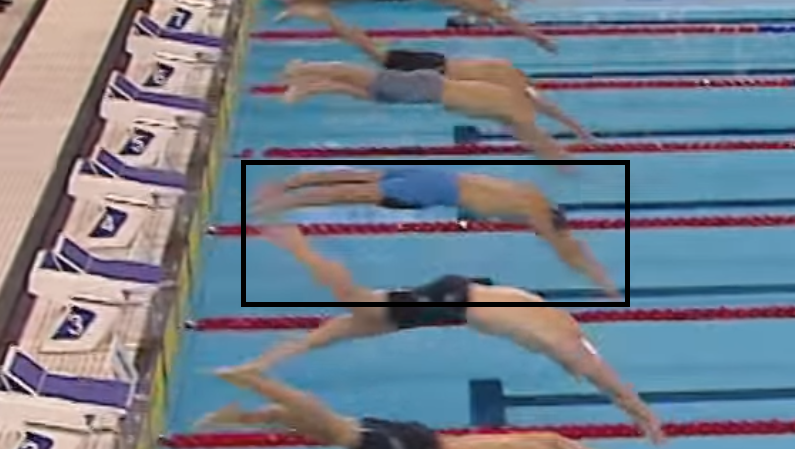}
        \captionsetup{justification=centering}
        \caption{Swimmers diving}
        \label{fig:swimming}
    \end{subfigure}
    ~
    \begin{subfigure}[t]{0.3\textwidth}
        \centering
        \includegraphics[height=1.2In]{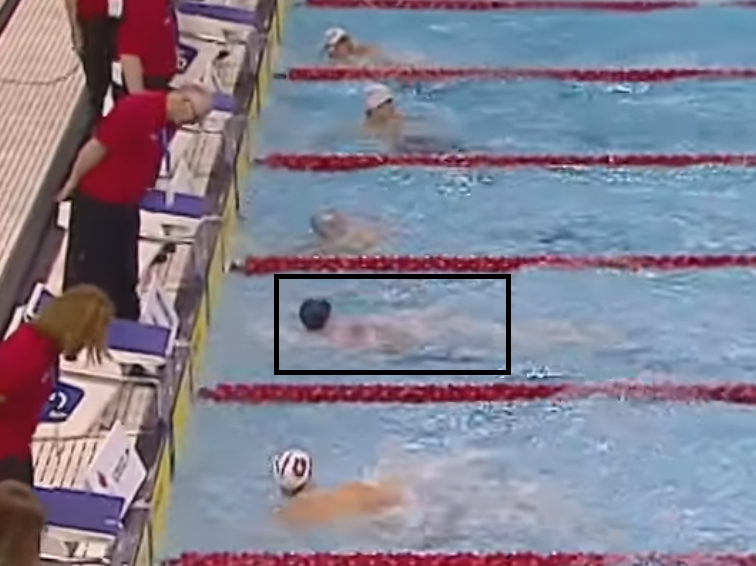}
        \captionsetup{justification=centering}
        \caption{Swimmers finishing}
        \label{fig:underwater}
        \end{subfigure}  
    
    \caption{Additional examples of swimmers states}
    
\end{figure*}

\section{Experiments Using One Pool}
When annotating video, many frames are highly correlated, and thus are redundant for training a detection or tracking algorithm. For this reason, tests were performed to reduce the amount of redundant annotations collected. In these experiments, we examined how to limit annotation to avoid collecting redundant data while still capturing sufficient data variety to train a successful model.   

\subsection{Collecting Swimmer Data for Tracking}
To illustrate the importance of efficiently and effectively collecting data, a simple and somewhat extreme example is considered. regard a single race video of a fifteen-hundred freestyle, at thirty frames per second, with eight swimmers, and with a length of sixteen minutes (regarded as a slow men's time one the world stage). The resulting video if annotated in full would result in more than 230,000 examples of swimmers, preforming all six classes of swimming in all its frames. When one considers the nature of a fifteen-hundred freestyle it is obvious these examples do not contain the right proportion of swimmer examples. This is evident as the examples will not contain all four strokes, their respective turns or the both genders of swimmers to say the least. There are other problems with collecting one single race and this is mentioned in the summary of swim race footage variability. Regardless, Using a custom-built labelling system, annotations of swimmers required an average of two seconds per bounding box. This means labelling the entirety of the aforementioned video would take over five days of continuous work. Because of the high redundancy in these images, such annotation would be an inefficient use of time. The following describes the experiments conducted in order to find a better annotation procedure.

\subsection{Extraction of Swimming Video Features}
The method used to test what frames in race video to annotate and what frames to skip was as follows. Using footage found on Swim USA's YouTube page \cite{swim_usa}, data was collected from a few videos of one competition, 2019 TYR Pro Swim Series - Bloomington. All possible strokes, turns and dives where present in the data collected. One in every three frames of the footage was annotated, as suggested in \cite{victor}. An exception was made with footage containing diving, in which case video was annotated frame by frame. This exception was due to the large amount of movement a dive contains and its short duration in time. The result was three-thousand frames of data with 25,000 examples of swimmers in various classes, the exact values can be seen in Table~\ref{tab:collected_data}.

\begin{table}
    \centering
    \begin{tabular}{l|r|r}
        Class & \# Annotations & Percent of Total\\
        \hline \hline
        ``On-blocks'' & 2,344 \hspace{0.45cm} & 10\% \hspace{0.5cm} \\
        Diving & 1,124 \hspace{0.45cm} & 5\%\hspace{0.6cm} \\
        Swimming & 13,009 \hspace{0.45cm} & 53\%\hspace{0.6cm} \\
        Underwater & 2,997 \hspace{0.45cm} & 12\%\hspace{0.6cm} \\
        Turning & 1,558 \hspace{0.45cm} & 6\%\hspace{0.6cm} \\
        Finishing & 3,534 \hspace{0.45cm} & 14\%\hspace{0.6cm} \\
        \hline
        Total & 24,566 \hspace{0.45cm} & 100\%\hspace{0.6cm} \\
    \end{tabular}
    \caption{The amount of collected data for each class}
    \label{tab:collected_data}
\end{table}

After the collection of this data, multiple models were trained with different subsets of this collected data to find the amount and distribution of data that produces optimal results. Optimal results being reducing the amount of redundancy in the dataset, while still obtaining detection results that are good enough. The first method of creating subsets was to randomly select a specified percentage of the three-thousand frames. The second method of subset creation was to randomly select a specified percentage of each class of the three-thousand frames. This method guaranteed that there will always be the same "percent of total" in all classes. Tests of the models using the second methods data will show if a certain class should have had more annotations in the initial collection phase.  

\section{Results} 
The Darknet-53, YOLOv3-416 model \cite{yolov3} and Darknet-15, YOLOv3-tiny-416 model \cite{tiny_yolo} was considered for testing. It was found that their results where almost identical and so the following tests were completed with the tiny model. In total, fifteen models where trained with 1\%, 2\%, 5\%, 10\%, 25\%, 50\%, 75\% and 100\% of the data collected. For each test the models where given the exact same architecture and parameters; the only way they differed was in the datasets they were trained on. Their performance was tested against a dataset of five-hundred frames that were not used in training. Half of the test set was obtained from the same pool used for training and the other half was obtained from a different pool but with similar conditions. These pools are designated as Bloomington and Winter National, respectively. The performance was gauged using mean average precision (mAP) for each class, for more details, see \cite{everingham}. The mAP of tracking was also collected. Tracking disregards the classes, the mAP value of tracking represents how good the model is at identifying the position of a swimmer in a pool. 

\begin{figure}[ht]
\centering
\includegraphics[width=8cm]{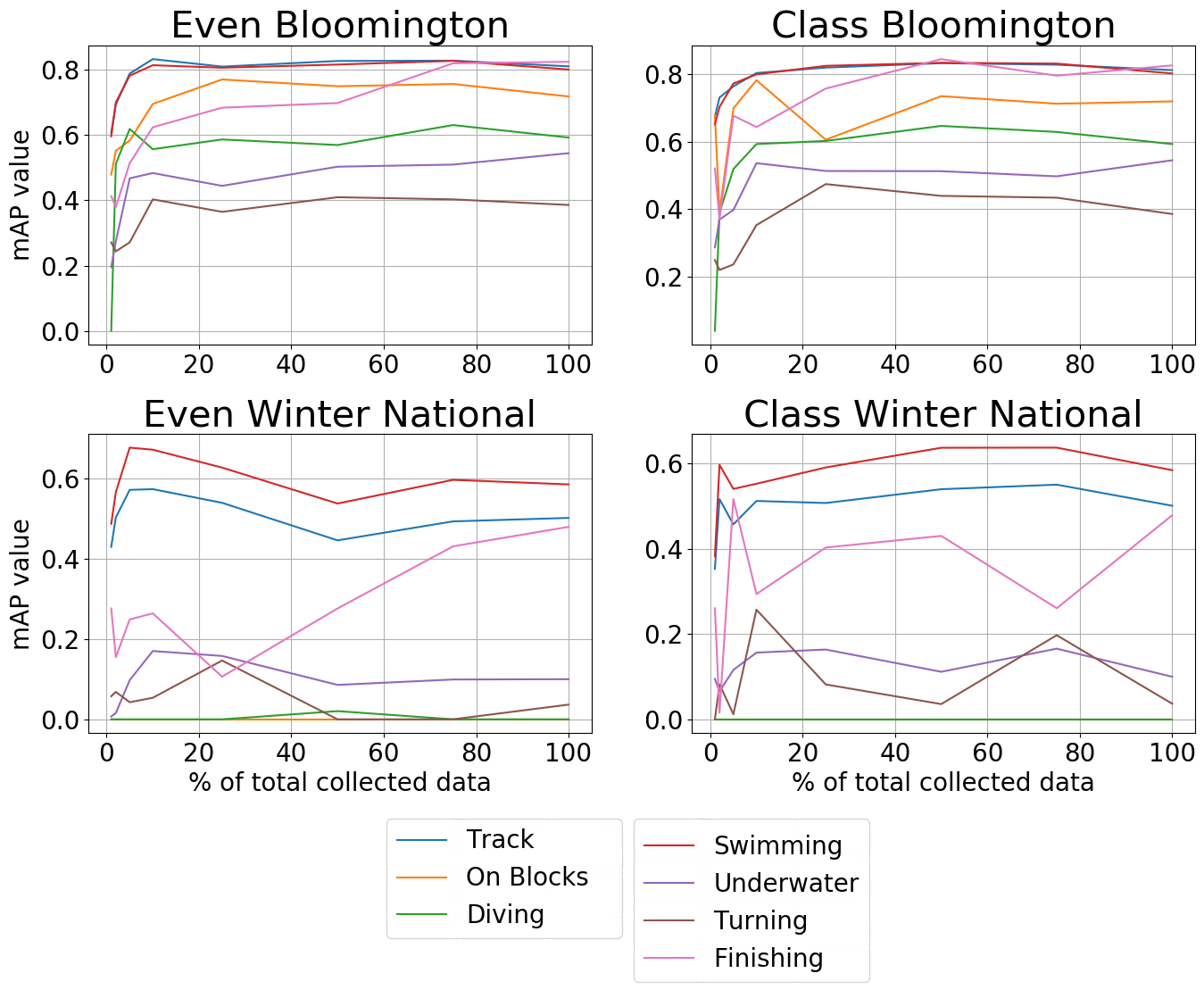}
\caption{Results from the data collection test}
\label{fig:res_plot}
\end{figure}

Figure \ref{fig:res_plot} shows a condensed breakdown of the results. The top plots represents testing of footage from the same pool, Bloomington. The bottom plots represents testing of footage from a different pool, Winter National. The x-axis represents the percent of data used for training and the y-axis represents the mAP value. Each line in a plot is either a different class or the tracking results. The plots on the left represent the first subset distribution and the plots on the right represent the second subset distribution. 

The first thing to notice from this test is that based on the top two graphs, using roughly twenty percent of the data collected was sufficient to produce comparable results. That is collecting data every fifteen frames and every five frames for diving. After reducing the data collected to less than twenty percent in a steep decline in overall performance is observed. 

\begin{figure}[ht]
\centering
\includegraphics[width=6cm]{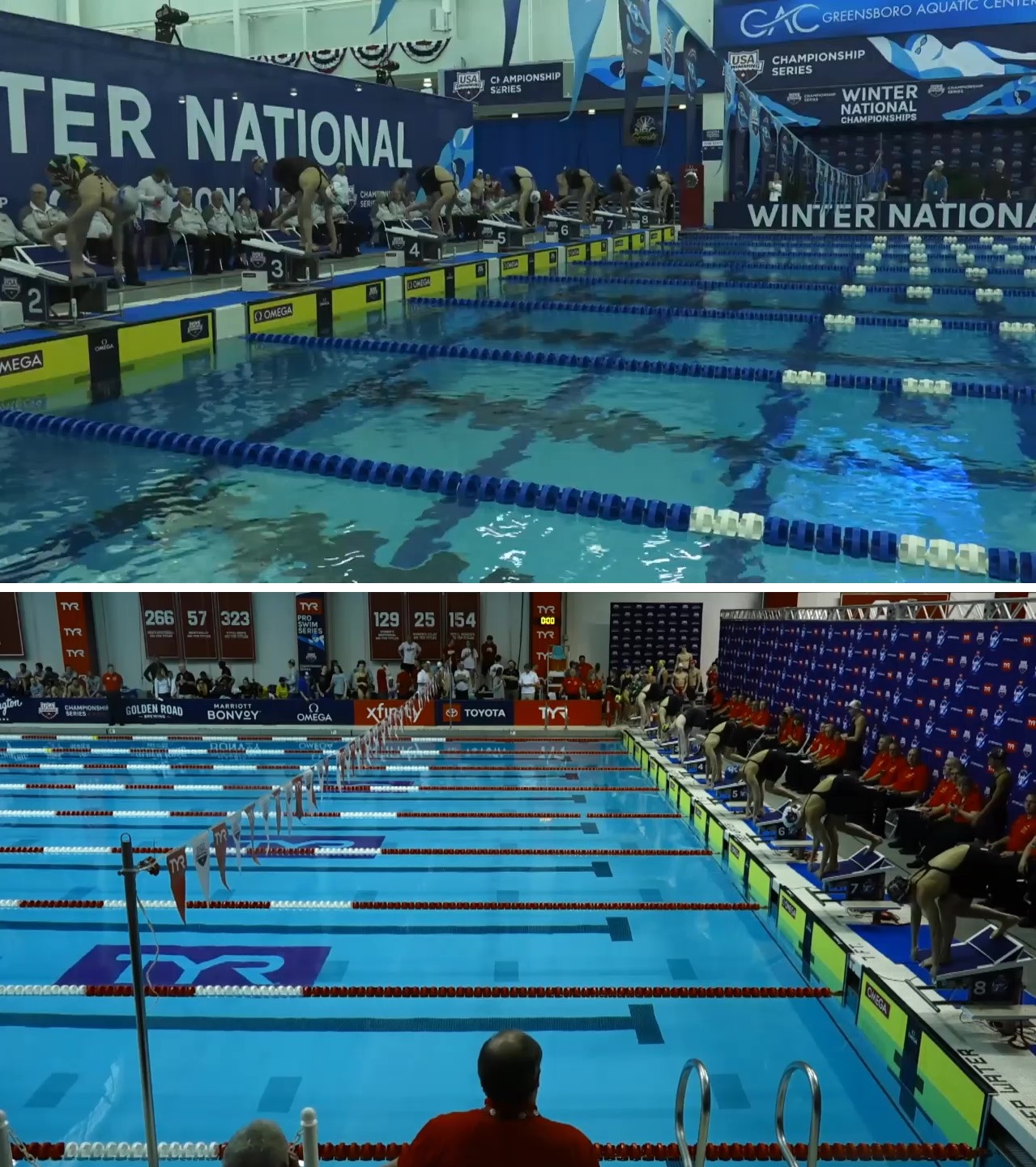}
\caption{Difference between dive camera angles: Winter National (top) and Bloomington (bottom)}
\label{fig:dive_comp}
\end{figure}

Next thing to notice is the extremely poor performance of the model when predicting diving at the Winter National pool and the less than optimal performance in the turning and underwater. These results are due to the difference in camera angles from one pool to the other, as can be seen in Figure \ref{fig:dive_comp}. This could have been partially fixed by flipping the training images horizontally but the camera angle between pools is different even with the horizontal flip. That being said, the swimming in Winter National was captured at roughly the same position as Bloomington. This is confirmed in figure \ref{fig:res_plot} as the Bloomington and Winter National swimming plots have the same profile. This is in contrast to the rest of the Winter National plots that have drastically different profiles than the Bloomington results.

Lastly, there seems to be no significant difference in the amount data for which mAP value sharply drops when comparing across all classes. This might indicate that collecting annotations once every fifteen frames (once every five for diving) is a good enough approximation. However, this conclusion is based on a relatively small test set. If more insight is to be gained on the distribution of classes collected from swimming footage, more tests need to be conducted.

\section{Summary}
In this paper we presented the first step in a project for the automation of swimming analytics. We began construction of a dataset through identifying the important aspects of data collection and annotation. The results suggest that data can be efficiently collected from a video by annotating frames at two or three frames a second (six frames a second for diving). Such analysis provided validation that under optimal circumstances a detection system can exist. Lastly, this experiment gave a general intuition of how deep learning detection models such as Darknet \cite{Redmon} respond to swimmer data. Specifically, a lighter detection model, such as Darknet-15 performs roughly the same as Darknet-53 for the detection of swimmers \cite{yolov3,tiny_yolo}. lastly, there is no reason to believe that a system such as this one should not work in a general sense, once given more training examples of swimmers in different competitions. 

With the tools put forth by this paper we are able to begin the next steps in automating swimming analytics. We will use the procedure presented here to collect more data from a variety of sources, creating an annotated dataset for swimming. Next, we will build better tracking solutions incorporating swimmer dynamics such as in~\cite{bewley2016simple}, and finally we will build metric collection solutions to automatically derive common swimming metrics such as stroke count and stroke length. The beauty of this work is that is very modular and as such can be built upon once the ground work has been completed. Upon completion, we are confident that this project will greatly help simplify the collection and use of swimming analytics, assisting coaches and athletes across all levels of swimming and even possibly help to increase viewer interest of swimming. 

\bibliography{Bibliography-File}
\bibliographystyle{aaai}
\end{document}